\documentclass[preprint,12pt]{morteza}
\usepackage{graphicx}
\usepackage{amssymb}
\usepackage{lineno}
\usepackage{times}
\usepackage{epsfig}
\usepackage{graphicx}
\usepackage{amsmath}
\usepackage{amssymb}
\usepackage{multirow}

\usepackage{graphicx}
\usepackage{bbold}
\usepackage{mathptmx}      
\usepackage{graphicx}
\usepackage{longtable}
\usepackage{booktabs}
\usepackage{subcaption}
\newtheorem{definition}{Definition}[section]

\journal{ }

\begin{document}

\begin{frontmatter}


\title{Scene Categorization from Contours: Medial Axis Based Salience Measures}

\author{Morteza Rezanejad$^{1}$, Gabriel Downs$^{1}$, John Wilder$^{2,3}$, Dirk B. Walther$^{3}$,\\Allan Jepson$^{2}$, Sven Dickinson$^{2}$,  and Kaleem Siddiqi$^{1}$}

\address{$^{1}$ School of Computer Science \& Centre for Intelligent Machines, McGill University, Montreal, QC, Canada, H3A 0E9\\
	$^{2}$ Department of Computer Science, University of Toronto, Toronto, ON, Canada, M5T 3A1\\
	$^{3}$ Department of Psychology, University of Toronto, Toronto, ON, Canada, M5S 3G3}

\begin{abstract}
The computer vision community has witnessed recent advances in scene categorization from images, with the state-of-the art systems now achieving impressive recognition rates on challenging benchmarks such as the Places365 dataset. Such systems have been trained on photographs which include color, texture and shading cues. The geometry of shapes and surfaces, as conveyed by scene contours, is not explicitly considered for this task.
Remarkably, humans can accurately recognize natural scenes from line drawings, which consist solely of contour-based shape cues. Here we report the first computer vision study on scene categorization of line drawings derived from popular databases including an artist scene
database, MIT67 and Places365. Specifically, we use off-the-shelf pre-trained CNNs to perform scene classification given only contour information as input, and find performance levels well above chance. We also show that medial-axis based contour salience methods can be used to select more informative subsets of contour pixels, and that the variation in CNN classification performance on various choices for these subsets is qualitatively similar to that observed in human performance. Moreover, when the salience measures are used to weight the contours, as opposed to pruning them, we find that these weights boost our CNN performance above that for unweighted contour input. That is, the medial axis based salience weights appear to add useful information that is not available when CNNs are trained to use contours alone.
\end{abstract}


\end{frontmatter}

\section{Introduction}
\label{intro}

Both biological and artificial vision systems are confronted with a potentially highly complex assortment of visual features in real-world scenarios. The features need to be sorted and grouped appropriately in order to support high-level visual reasoning, including the recognition or categorization of objects or entire scenes. In fact, scene categorization cannot be easily disentangled from the recognition of objects, since scene classes are often defined by a collection of objects in context. A beach scene, for example, would typically contain umbrellas, beach chairs and people in bathing suits, all of whom are situated next to a body of water. A street scene might have roads with cars, cyclists and pedestrians as well as buildings along the edge. How might computer vision systems tackle this problem of organizing input visual features to support 
scene categorization?

In human vision, perceptual organization is thought to be effected by a set of heuristic grouping rules originating from Gestalt psychology \cite{koffka1922perception}. Such rules posit that visual elements ought to be grouped together if they are, for instance,  similar in appearance, in close proximity, or if they are symmetric or parallel to each other.  Developed on an ad-hoc, heuristic basis originally, these rules have been validated empirically, even though their precise neural mechanisms remain elusive. Grouping cues, such as those based on symmetry, are thought to aid in high-level visual tasks such as object detection, because symmetric contours are more likely to be caused by the projection of a symmetric object than to occur accidentally. In the categorization of complex real-world scenes by human observers, local contour symmetry does indeed provide a perceptual advantage \cite{wilder2019local}, but the connection to the recognition of individual objects is not as straightforward as it may appear. 

\begin{figure*}[!ht]
	\begin{tabular}{@{\hskip0pt}c@{\hskip2pt} c @{\hskip2pt}c@{\hskip0pt}}
		\fbox{\includegraphics[width=0.31\textwidth]{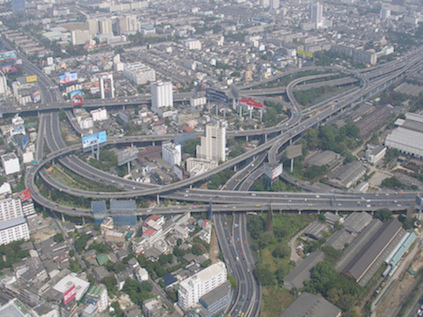}} &
		\fbox{\includegraphics[width=0.31\textwidth]{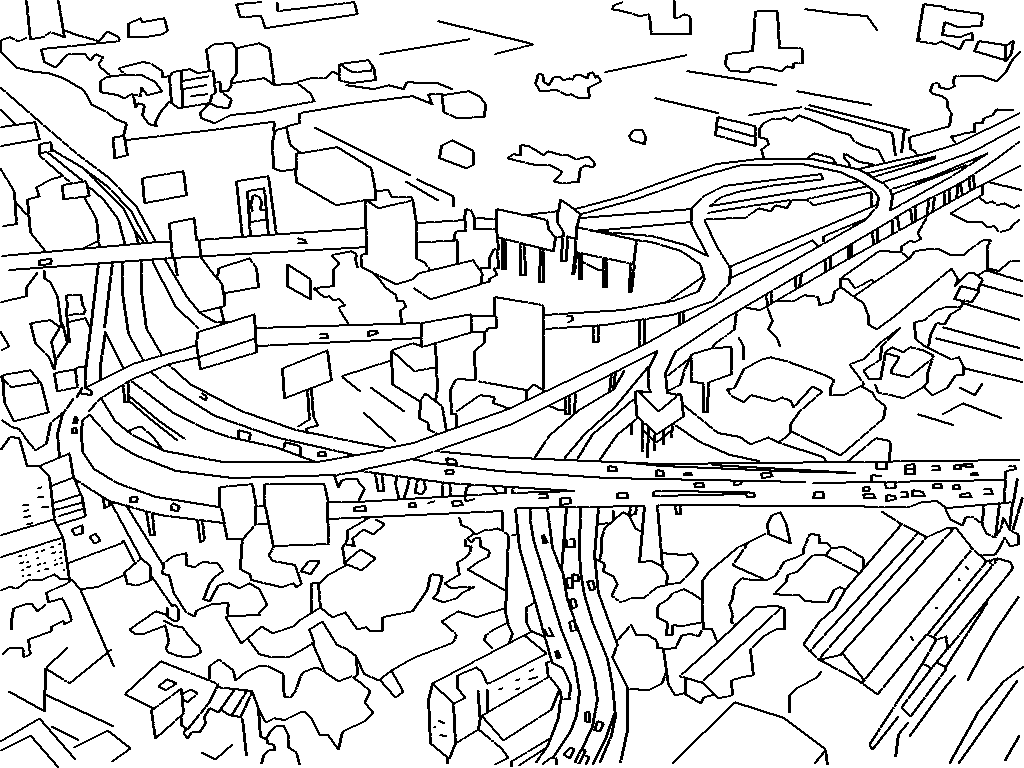}} &
		\fbox{\includegraphics[width=0.31\textwidth]{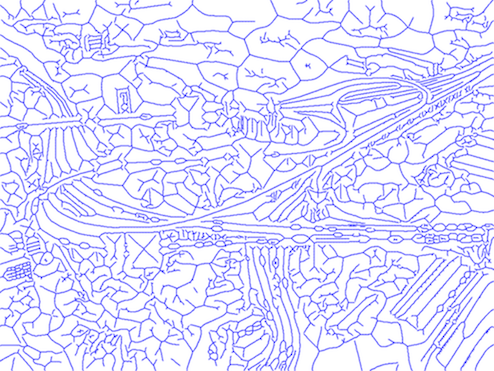}}\\
		{\small Photograph} & {\small Line Drawing} & {\small AOF Medial Axes}\\
		\fbox{\includegraphics[width=0.31\textwidth]{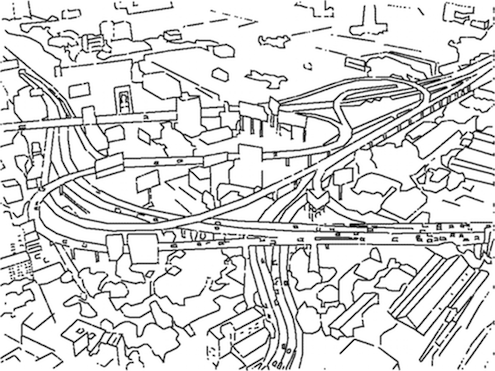}} &			\fbox{\includegraphics[width=0.31\textwidth]{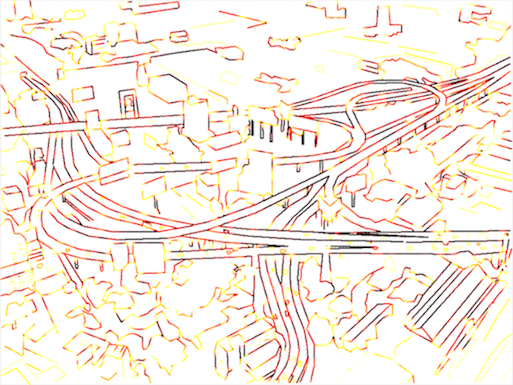}} &
		\fbox{\includegraphics[width=0.31\textwidth]{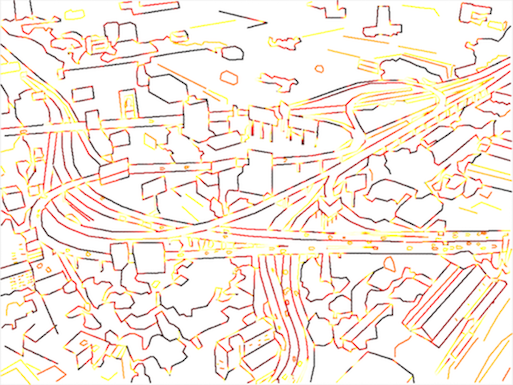}}\\
		{\small Reconstruction} & {\small Symmetry}\hspace{3mm}\includegraphics[width=0.14\textwidth , trim=0 0 0 -1]{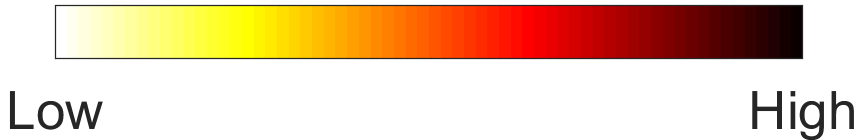} & {\small Separation} \hspace{3mm}\includegraphics[width=0.14\textwidth, trim=0 0 0 -1]{hot_colormap.PNG}
	\end{tabular}
	\caption{(Best viewed by zooming in on the PDF.) An illustration of our approach on an example from a database of line drawings by artists of photographs of natural scenes. The bottom left panel shows the reconstruction of the artist-generated line drawing from the AOF medial axes. To its right we present a hot colormap visualization of two of our medial axis based contour salience measures.}
	\label{fig:overview}
\end{figure*}
In computer vision, symmetry, proximity, good continuation, contour closure and other cues have been used for image segmentation, curve inference, object recognition, object manipulation, and other tasks \cite{marr1978representation, biederman1987recognition, elder1996computing,sarker1999perceptual}. Instantiations of such organizational principles have found their way into many computer vision algorithms and have been the subject of regular workshops on perceptual organization in artificial vision systems. However, perceptually motivated salience measures to facilitate scene categorization have received little attention thus far. This may be a result of the ability of CNN based systems to accomplish scene categorization on challenging databases, in the presence of sufficient training data, directly from pixel intensity and colour in photographs
\cite{sharif2014cnn,Szegedy_2016_CVPR,hoo2016deep,zhou2018places}. CNNs begin by extracting simple features, including oriented edges, which are then successively combined into more and more complex features in a succession of convolution, nonlinear activation and pooling operations. The final levels of CNNs are typically fully connected, which enable learning of object or scene categories \cite{song2015sun,bai2017growing,girshick2014rich,ren2015faster}. Unfortunately, present CNN architectures do not explicitly allow for  properties of object shape to be elucidated. Human observers, in contrast, recognize an object's shape as an inextricable aspect of its properties, along with its category or identity \cite{kellman1991theory}.

Comparisons between CNNs and human and monkey neurophysiology appear to indicate that CNNs replicate the entire visual hierarchy \cite{Guclu2005, Cadieu2014}. Does this mean that the problem of perceptual organization is now irrelevant for computer vision? 
In the present article we argue that this is not the case. Rather, we show that CNN-based scene categorization systems, just like human observers, can benefit from explicitly computed contour measures derived from Gestalt grouping cues. We here demonstrate the computation of these measures as well as their power to aid in the categorization of complex real-world scenes.


To effect our study, with its focus on the geometry of scene contours, we choose to use the medial axis transform (MAT) as a representation. We apply a robust algorithm for computing the medial axes to analyze line drawings of scenes of increasing complexity. The algorithm uses the average outward flux of the gradient of the Euclidean distance function through shrinking circular disks \cite{dimitrov2003flux}. With its explicit representation of the regions between scene contours, the medial axis allows us to directly capture salience measures related to local contour separation and local contour symmetry. We introduce two novel measures of local symmetry using ratios of length functions derived from the medial axis radius along skeletal segments. As ratios of commensurate quantities, these are unitless measures, which are therefore invariant to image re-sizing. We describe methods of computing our perceptually motivated salience measures from line drawings of photographs of complex real-world scenes, covering databases of increasing complexity. Figure \ref{fig:overview} presents an illustrative example of a photograph from an artist scenes database, along with two of our medial axes based contour salience maps. Observe how the ribbon symmetry based measure highlights the boundaries of highways. Our experiments show that scene contours weighted by these measures can boost CNN-based scene categorization accuracy, despite the absence of colour, texture and shading cues. Our work indicates that measures of contour grouping, that are simply functions of the contours themselves, are beneficial for scene categorization by computers, yet that they are not automatically extracted by state-of-the-art CNN-based scene recognition systems. The critical remaining question is whether this omission is due to the CNN architecture being unable to model these weights or whether this has to do with the (relatively standard) training regime. We leave this for further study.

\section{Average Outward Flux Based Medial Axes}
\label{sec:AOF}
\label{medial_axis}

In Blum's grassfire analogy the medial axis is associated with the quench points of a fire that is lit at the boundary of a field of grass \cite{Blum1973-za}. In the present paper, that boundary is the set of scene contours, and the field of grass is the space between them. An equivalent notion of the medial axis is that of the locus of centres of maximal inscribed disks in the region between scene contours, along with the radii of these disks. The geometry and methods for computing the medial axis that we leverage are based on a notion of average outward flux, as discussed in further detail below. We apply the same algorithm to each distinct connected region between scene contours. These regions are obtained by morphological operations to decompose the original line drawing.

\begin{definition}
	Assume an $n$-dimensional open connected region $\Omega$, with its boundary given by $\partial \Omega \in \mathbb{R}^n$ such that $\bar{\Omega} = \Omega \cup \partial \Omega$. An open disk $D \in \mathbb{R}^n $ is a maximal inscribed disk in $\bar{\Omega}$ if $D \subseteq \bar{\Omega}$ but for any open disk $D'$ such that $D \subset D'$, the relationship $D' \subseteq \bar{\Omega}$ does not hold.
\end{definition}

\begin{definition}
	The \textit{Blum medial locus} or \textit{skeleton}, denoted by $Sk(\Omega)$, is the locus of centers of all maximal inscribed disks in $\partial \Omega$.
\end{definition}

Topologically, $Sk(\Omega)$ consists of a set of branches, about which the scene contours are locally mirror symmetric, that join at branch points to form the complete skeleton.  A skeletal branch is a set of contiguous regular points from the skeleton that lie between a pair of junction points, a pair of end points or an end point and a junction point. At regular points the maximal inscribed disk touches the boundary at two distinct points.
As shown by Dimitrov \textit{et al.} \cite{dimitrov2003flux} medial axis points can be analyzed by considering the behavior of the average outward flux (AOF) of the gradient of the Euclidean distance function through the boundary of the connected region. Let $R$ be the region with boundary $\partial R$, and let $\mathbf{N}$ be the outward normal at each point on the boundary  $\partial R$. The AOF is given by the limiting value of $\frac{\int_{\partial R} \langle\dot{\mathbf{q}},\mathbf{N}\rangle ds}{\int_{\partial R} ds}$, as the region is shrunk. Here $\dot{\mathbf{q}} = \nabla \mathbf{D}$, with $\mathbf{D}$ the Euclidean distance function to the connected region's boundary, and the limiting behavior is shown to be different for each of three cases: regular points, branch points and end points. When the region considered is a shrinking disk, at regular points of the medial axis the AOF works out to be $- \frac{2}{\pi}\sin\theta$, where $\theta$ is the object angle, the acute angle that a vector from a skeletal point to the point where the inscribed disk touches the boundary on either side of the medial axis makes with the tangent to the medial axis. This quantity is negative because it is the inward flux that is positive. Furthermore, the limiting $AOF$ value for all points not located on the medial axis is zero.

This provides a foundation for both computing the medial axis for scene contours and for mapping the computed medial axis back to them. First, given the Euclidean distance function from scene contours, one computes the limiting value of the AOF through a disk of shrinking radius and associates locations where this value is non-zero with medial axis points (Figure \ref{fig:overview}, top right). Then, given the AOF value at a regular medial axis point, and an estimate of the tangent to the medial axis at it, one rotates the tangent by $\pm \theta$ and then extends a vector out on either side by an amount given by the radius function, to reconstruct the boundary (Figure \ref{fig:overview}, bottom left). In our implementations we discretize these computations on a fine grid, along with a dense sampling of the boundary of the shrinking disk, to get high quality scene contour representation.

\section{Medial Axis Based Contour Saliency}
\label{sec:salience}
Owing to the continuous mapping between the medial axes and scene contours, the medial axes provide a convenient representation for designing and computing Gestalt contour salience measures based on local contour separation and local symmetry. 
A measure to reflect local contour separation can be designed using the radius function along the medial axis, since this measure gives the distance to the two nearest scene contours on either side.
Local parallelism between scene contours, or ribbon symmetry, can also be directly captured by examining the degree to which the radius function along the medial axis between them remains locally constant. Finally, if taper is to be allowed between contours, as in the case of a set of railway tracks extending to the horizon under perspective projection, one can examine the degree to which the first derivative of the radius function is constant along a skeletal segment. We introduce novel measures to capture local separation, ribbon symmetry and taper, based on these ideas.

In the following we shall let $p$ be a parameter that runs along a medial axis segment, $\mathbf{C}(p)=(x(p),y(p))$ be the coordinates of points along that segment, and $R(p)$ be the medial axis radius at each point. We shall consider the interval $p \in [\alpha,\beta]$ for a particular medial segment. The arc length of that segment is given by 
\begin{equation}
	\label{eq:length_function}
	L= \int_{\alpha} ^{\beta} || \frac{\partial \mathbf{C}}{\partial p} || dp = \int_{\alpha} ^{\beta}  (x_p^2+y_p^2)^{\frac{1}{2}}dp.
\end{equation}	

\subsection{Separation Salience}
We now introduce a salience measure based on the local separation between two scene contours associated with the same medial axis segment. Consider the interval $p \in [\alpha,\beta]$. With $R(p) > 1$ in pixel units (because two scene contours cannot touch) we introduce the following contour separation based salience measure:
\begin{equation}
	S_{Separation} = 1-\Big(\int_\alpha ^ \beta \frac{1}{R(p)} dp\Big)/(\beta-\alpha).
\end{equation}
This quantity falls in the interval $[0,1]$. The measure increases with increasing spatial separation between the two contours. In other words, scene contours that exhibit further (local) separation are more salient by this measure.


\subsection{Ribbon Symmetry Salience}
\label{subsec:ribbon}
Now consider the curve ${\Psi} = (x(p),y(p),R(p))$. Similar to Equation \ref{eq:length_function}, the arc length of $\Psi$ is computed as:
\begin{equation}
	\label{eq:psi_r}
	L_\Psi= \int_{\alpha} ^{\beta} || \frac{\partial {\Psi}}{\partial p} || dp = \int_{\alpha} ^{\beta}  (x_p^2+y_p^2+R_p^2)^{\frac{1}{2}}dp.
\end{equation}	
When two scene contours are close to being parallel locally, $R(p)$ will vary slowly along the medial segment. This motivates the following ribbon symmetry salience measure:
\begin{equation}
	S_{Ribbon} = \frac{ L}{L_\Psi} = \frac{\int_{\alpha} ^{\beta}  (x_p^2+y_p^2)^{\frac{1}{2}}dp }{\int_{\alpha} ^{\beta}  (x_p^2+y_p^2+R_p^2)^{\frac{1}{2}}dp}.
\end{equation}
This quantity also falls in the interval $[0,1]$ and is invariant to image scaling since the integral involves a ratio of unitless quantities. The measure is designed to increase as the scene contours on either side become more parallel, such as the two sides of a ribbon.

\subsection{Taper Symmetry Salience}
A notion that is closely related to that of ribbon symmetry is taper symmetry; two scene contours are taper symmetric when the medial axis between them has a radius function that is changing at a constant rate, such as the edges of two parallel contours in 3D when viewed in perspective. To capture this notion of symmetry, we introduce a slight variation where we consider a type of arc-length of a curve $\Psi' = (x(p),y(p),\frac{dR(p)}{dp})$. Specifically, we introduce the following taper symmetry salience measure:
\begin{equation}
	\label{eq:actualddR}
	S_{Taper} = \frac{L}{L_{\Psi'}} = \frac{\int_{\alpha} ^{\beta}  (x_p^2+y_p^2)^{\frac{1}{2}}dp }{\int_{\alpha} ^{\beta}  (x_p^2+y_p^2+(RR_{pp})^2)^{\frac{1}{2}}dp}.
\end{equation}
The bottom integral is not exactly an arc-length, due to the multiplication of $R_{pp}$ by the factor $R$. This modification is necessary to make the overall ratio unitless. This quantity also falls in the interval $[0,1]$ and is invariant to image scaling. The measure is designed to increase as the scene contours on either side become more taper symmetric, as in the shape of a funnel, or the sides of a railway track.
\begin{figure}[ht]
	\centering
	\begin{tabular}{c@{\hskip3pt}c@{\hskip3pt}c}
		\hline
		& Shape &\\
		\hline
		\includegraphics[width=0.2\textwidth, trim=0 0 0 -5]{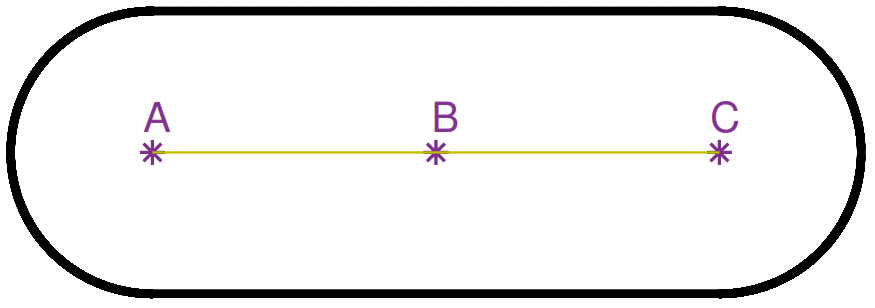}&
		\includegraphics[width=0.2\textwidth, trim=0 0 0 -10]{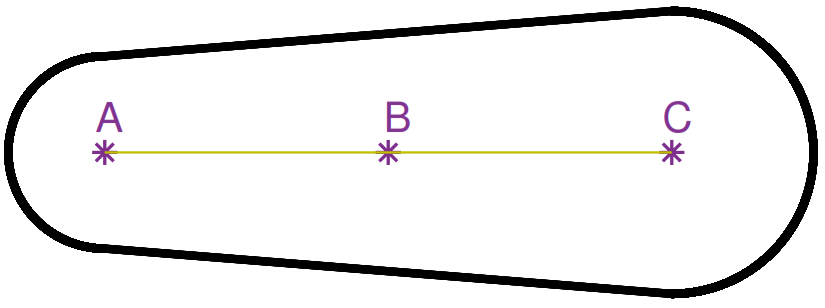}&
		\includegraphics[width=0.2\textwidth]{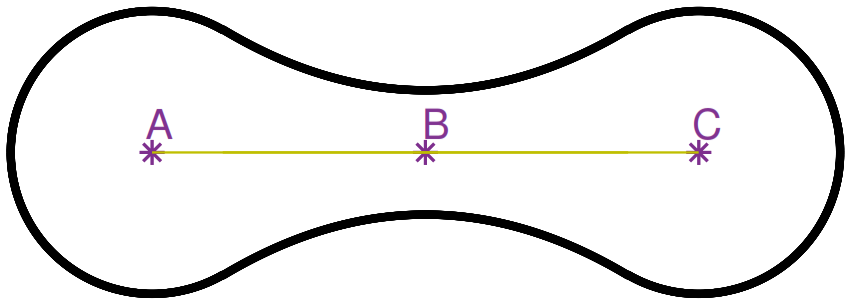}\\
		\hline
		& Ribbon Salience & \\
		\hline
		\includegraphics[width=0.2\textwidth, trim=0 0 0 -10]{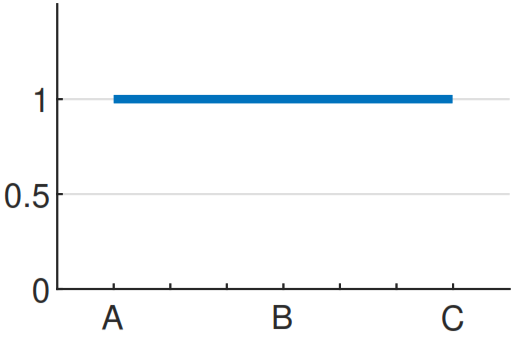}&
		\includegraphics[width=0.2\textwidth]{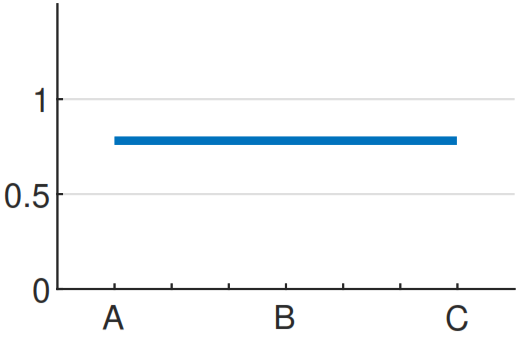}&
		\includegraphics[width=0.2\textwidth]{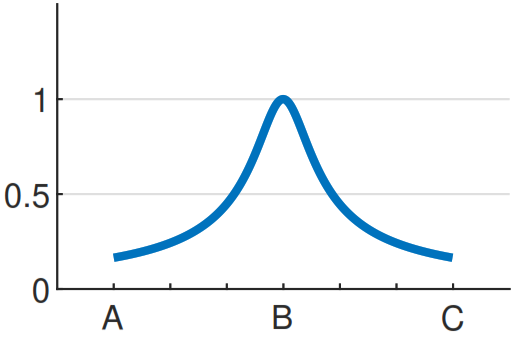}\\
		\hline
		& Taper Salience & \\
		\hline
		\includegraphics[width=0.2\textwidth, trim=0 0 0 -10]{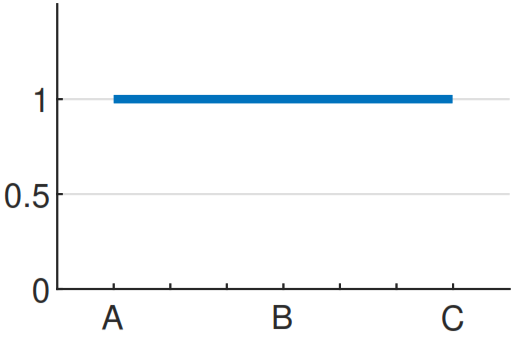}&
		\includegraphics[width=0.2\textwidth]{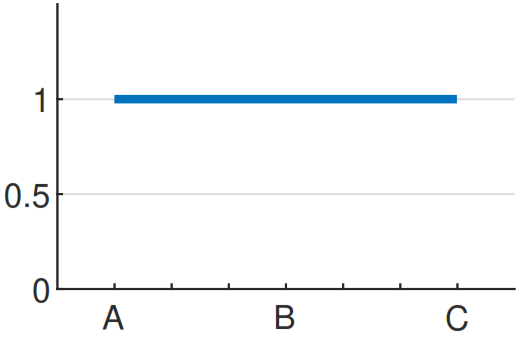}&
		\includegraphics[width=0.2\textwidth]{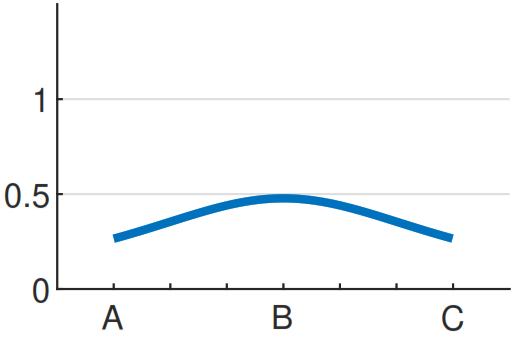}\\
		\hline
		& Separation Salience &\\
		\hline
		\includegraphics[width=0.2\textwidth, trim=0 0 0 -10]{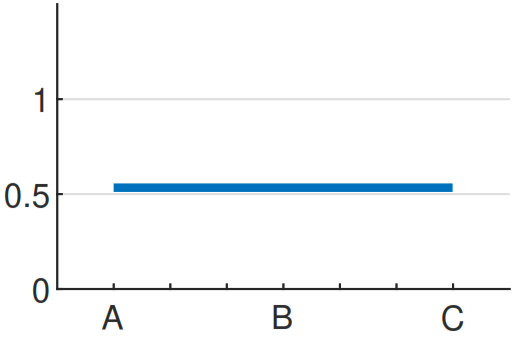}&
		\includegraphics[width=0.2\textwidth]{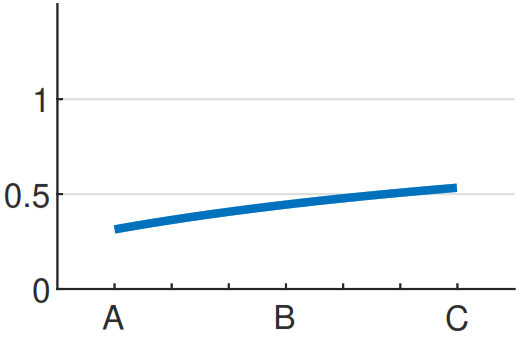}&
		\includegraphics[width=0.2\textwidth]{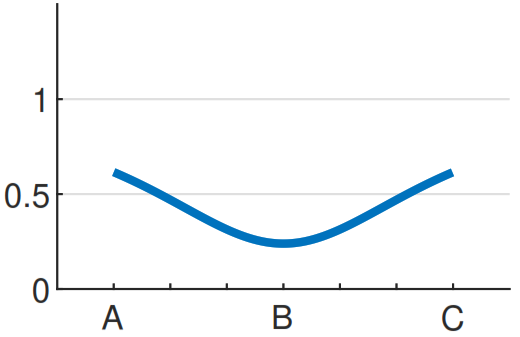}\\
		\hline
	\end{tabular}
	\caption{An illustration of ribbon symmetry salience, taper symmetry salience and contour separation salience for three different contour configurations. See text for a discussion. These measures are all invariant to 2D similarity transforms of the input contours}
	\label{fig:methods_examples}
\end{figure}

To gain an intuition behind these perceptually driven contour salience measures, we provide three illustrative examples in Fig. \ref{fig:methods_examples}. The measures are not computed point-wise, but rather for a small interval $[\alpha,\beta]$ centered at each medial axis point (see Section \ref{subsec:computing} for details). When the contours are parallel all three measures are constant along the medial axis (left column). The middle figure has high taper symmetry but lower ribbon symmetry, with contour separation salience increasing from left to right. Finally, for the dumbbell shape, all three measures vary (third column).


\section{Experiments and Results}
\label{sec:experiments}

\subsection{Artist Generated Line Drawings}

\textbf{Artist Scenes Database}:
Color photographs of six categories of natural scenes (beaches, city streets, forests, highways, mountains, and offices) were downloaded from the internet, and those rated as the best exemplars of their respective categories by workers on Amazon Mechanical Turk were selected. Line drawings of these photographs were generated by trained artists at the Lotus Hill Research Institute  \cite{walther2011simple}. Artists traced the most important and salient lines in the photographs on a graphics tablet using a custom graphical user interface. Contours were saved as successions of anchor points. For the experiments in the present paper, line drawings were rendered by connecting anchor points with straight black lines on a white background at a resolution of $1024 \times 768$ pixels. The resulting database had 475 line drawings in total with 79 exemplars from each of 6 categories: beaches, mountains, forests, highway scenes, city scenes and office scenes.

\subsection{Machine Generated Line Drawings}
\begin{figure}[!ht]
	
	\begin{tabular}{c@{\hskip3pt} c @{\hskip3pt}c}
		\fbox{\includegraphics[width=0.3\textwidth]{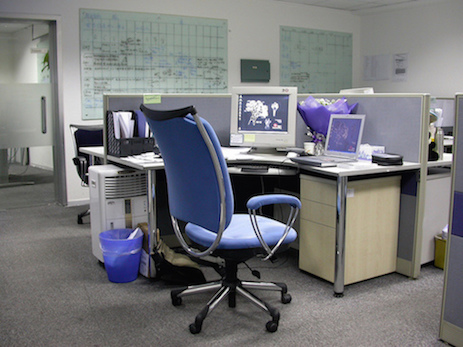}} &
		\fbox{\includegraphics[width=0.3\textwidth]{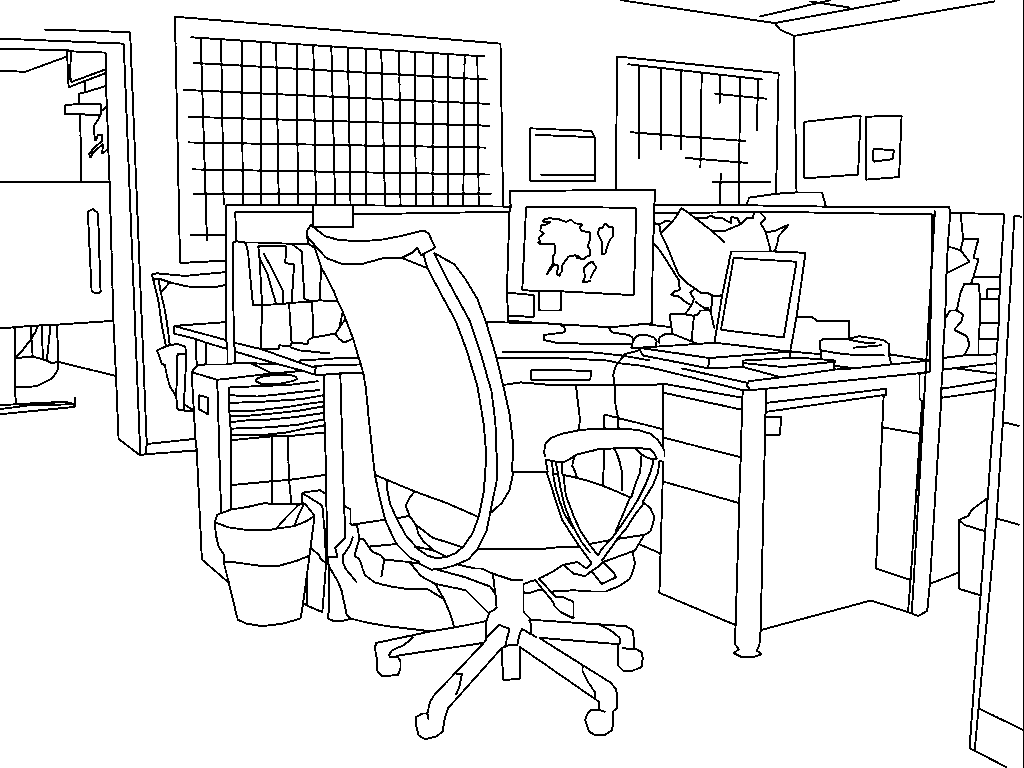}} &
		\fbox{\includegraphics[width=0.308\textwidth]{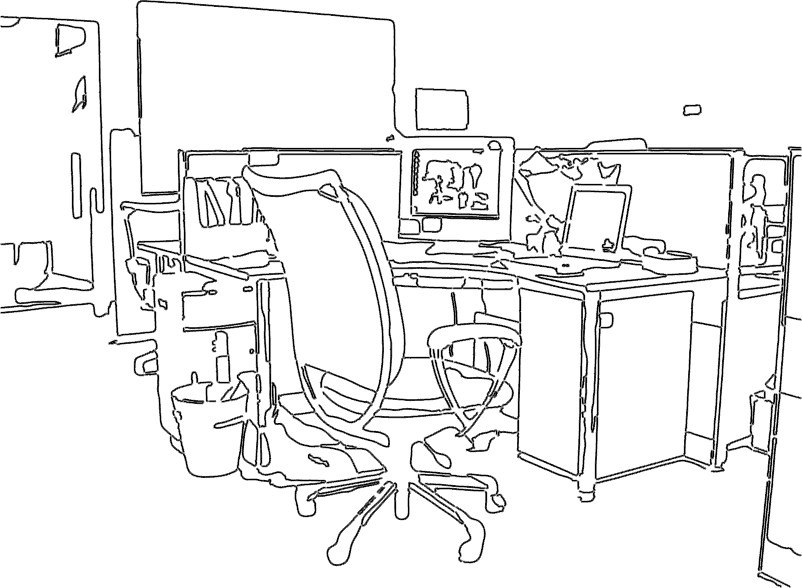}}\\
		Photograph & Artist & Machine
	\end{tabular}
	\caption{(Best viewed by zooming in on the PDF.) A comparison between a machine-generated line drawing and one drawn by an artist, for an office scene from the Artist Scenes database.}
	\label{fig:machine_generated}
	
\end{figure}

\begin{figure*}[!t]
	\begin{tabular}{c@{\hskip10pt}c@{\hskip2pt}c@{\hskip8pt}c@{\hskip2pt}c@{\hskip8pt}c@{\hskip2pt}c}
		
		\parbox[t]{2mm}{\multirow{3}{*}{\rotatebox[origin=c]{90}{\Large Places365}}}&
		\fbox{\includegraphics[width=18mm,height = 12mm]{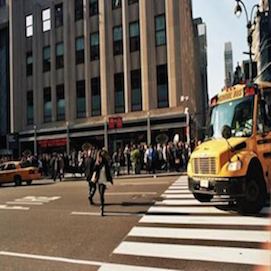}}&
		\fbox{\includegraphics[width=18mm,height = 12mm]{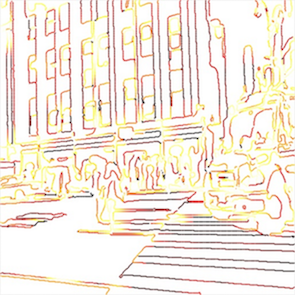}}&
		\fbox{\includegraphics[width=18mm,height = 12mm]{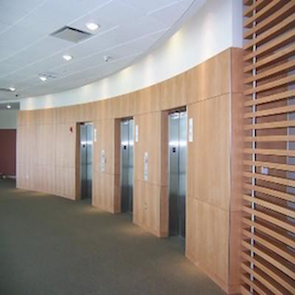}}&
		\fbox{\includegraphics[width=18mm,height = 12mm]{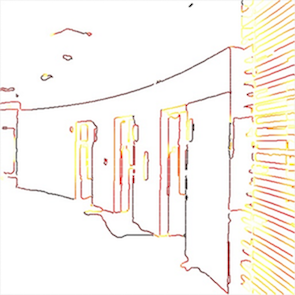}}&
		\fbox{\includegraphics[width=18mm,height = 12mm]{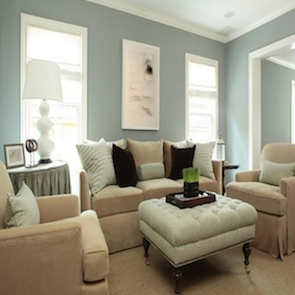}}&
		\fbox{\includegraphics[width=18mm,height = 12mm]{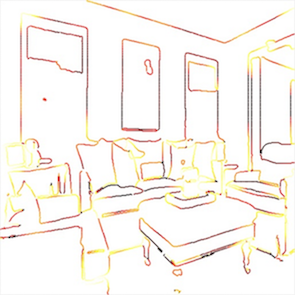}}\\
		& \fbox{\includegraphics[width=18mm,height = 12mm]{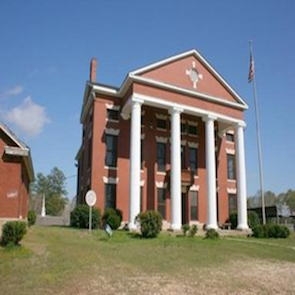}}&
		\fbox{\includegraphics[width=18mm,height = 12mm]{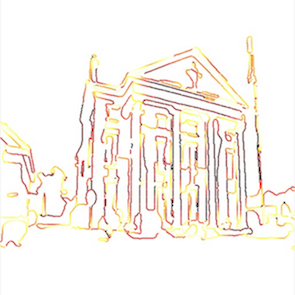}}&
		\fbox{\includegraphics[width=18mm,height = 12mm]{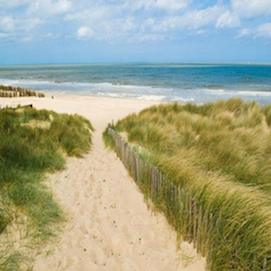}}&
		\fbox{\includegraphics[width=18mm,height = 12mm]{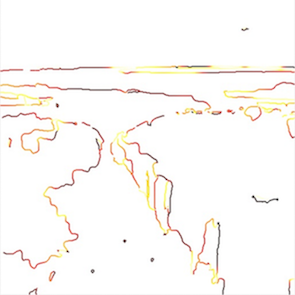}}&
		\fbox{\includegraphics[width=18mm,height = 12mm]{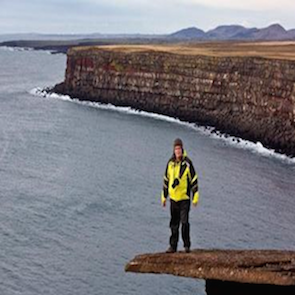}}&
		\fbox{\includegraphics[width=18mm,height = 12mm]{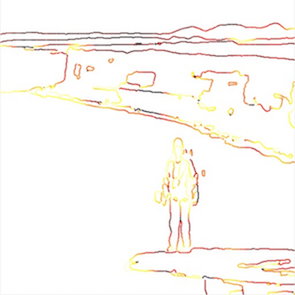}}\\\vspace{2mm}   
		& \fbox{\includegraphics[width=18mm,height = 12mm]{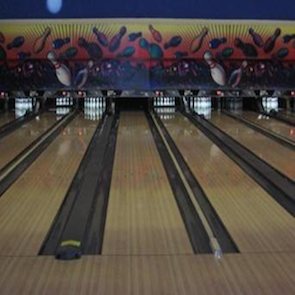}}&
		\fbox{\includegraphics[width=18mm,height = 12mm]{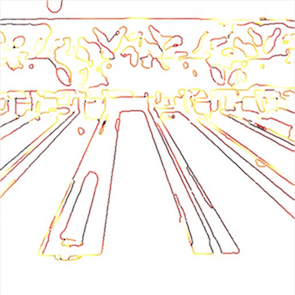}}&
		\fbox{\includegraphics[width=18mm,height = 12mm]{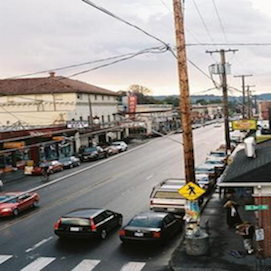}}&
		\fbox{\includegraphics[width=18mm,height = 12mm]{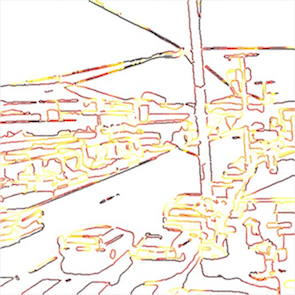}}&
		\fbox{\includegraphics[width=18mm,height = 12mm]{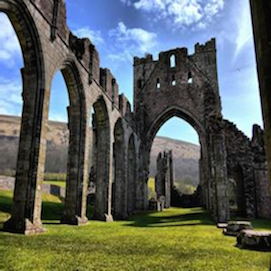}}&
		\fbox{\includegraphics[width=18mm,height = 12mm]{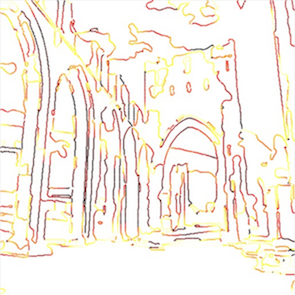}}\\
		
		\parbox[t]{2mm}{\multirow{3}{*}{\rotatebox[origin=c]{90}{\Large MIT67}}}&
		\fbox{\includegraphics[width=18mm,height = 12mm]{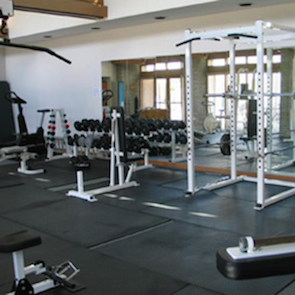}}&
		\fbox{\includegraphics[width=18mm,height = 12mm]{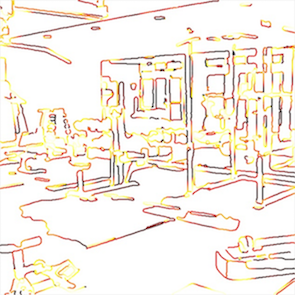}}&
		\fbox{\includegraphics[width=18mm,height = 12mm]{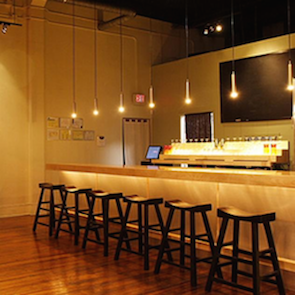}}&
		\fbox{\includegraphics[width=18mm,height = 12mm]{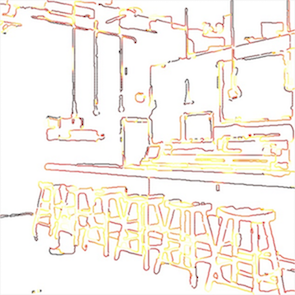}}&
		\fbox{\includegraphics[width=18mm,height = 12mm]{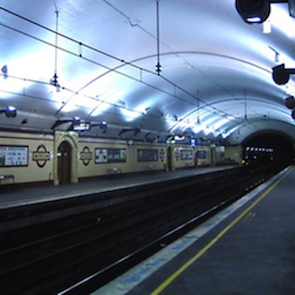}}&
		\fbox{\includegraphics[width=18mm,height = 12mm]{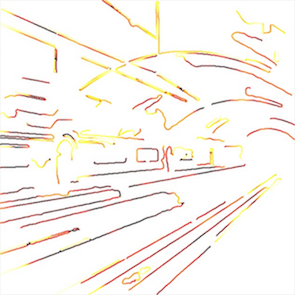}}\\
		&\fbox{\includegraphics[width=18mm,height = 12mm]{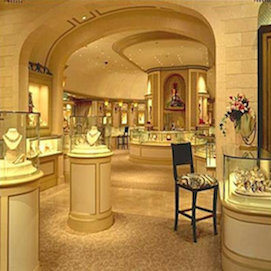}}&
		\fbox{\includegraphics[width=18mm,height = 12mm]{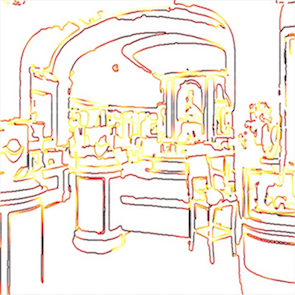}}&
		\fbox{\includegraphics[width=18mm,height = 12mm]{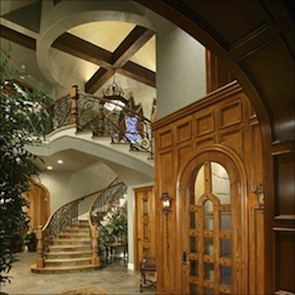}}&
		\fbox{\includegraphics[width=18mm,height = 12mm]{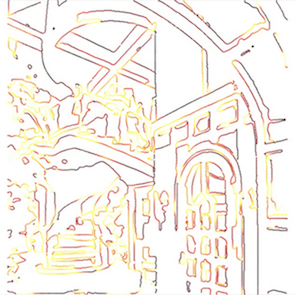}}&
		\fbox{\includegraphics[width=18mm,height = 12mm]{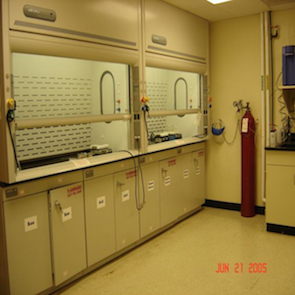}}&
		\fbox{\includegraphics[width=18mm,height = 12mm]{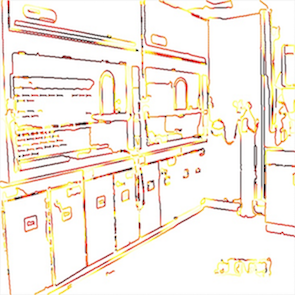}}\\\vspace{2mm}
		&\fbox{\includegraphics[width=18mm,height = 12mm]{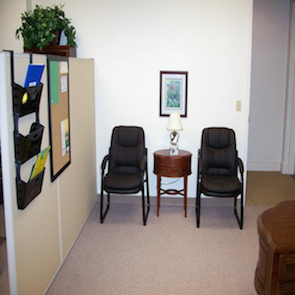}}&
		\fbox{\includegraphics[width=18mm,height = 12mm]{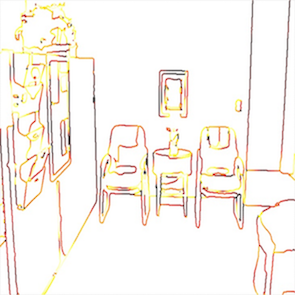}}&
		\fbox{\includegraphics[width=18mm,height = 12mm]{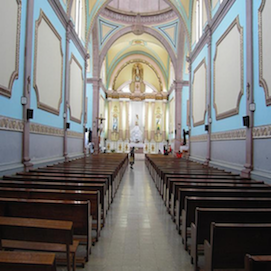}}&
		\fbox{\includegraphics[width=18mm,height = 12mm]{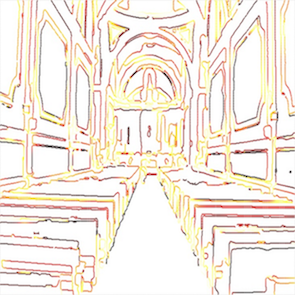}}&
		\fbox{\includegraphics[width=18mm,height = 12mm]{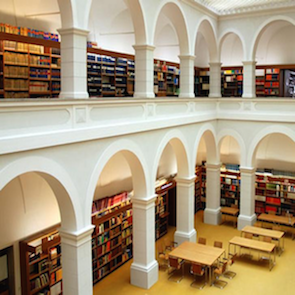}}&
		\fbox{\includegraphics[width=18mm,height = 12mm]{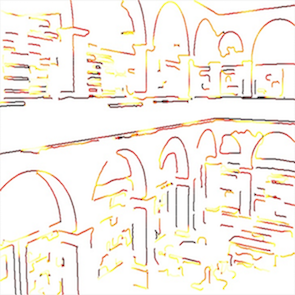}}\\
		
		\parbox[t]{2mm}{\multirow{3}{*}{\rotatebox[origin=c]{90}{\Large Artist Scenes}}}& 
		\fbox{\includegraphics[width=18mm,height = 12mm]{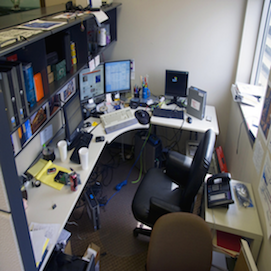}}&
		\fbox{\includegraphics[width=18mm,height = 12mm]{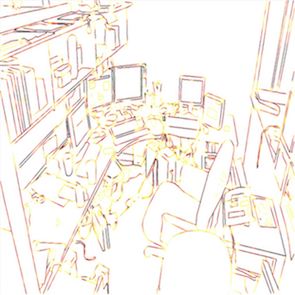}}&
		\fbox{\includegraphics[width=18mm,height = 12mm]{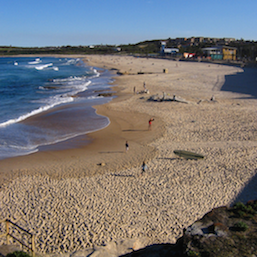}}&
		\fbox{\includegraphics[width=18mm,height = 12mm]{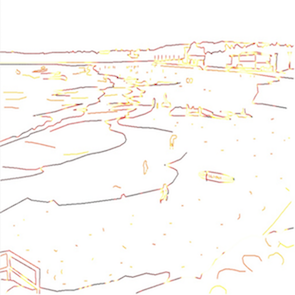}}&
		\fbox{\includegraphics[width=18mm,height = 12mm]{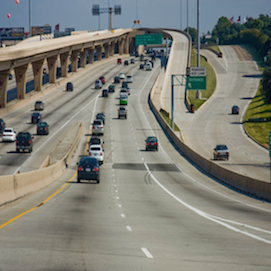}}&
		\fbox{\includegraphics[width=18mm,height = 12mm]{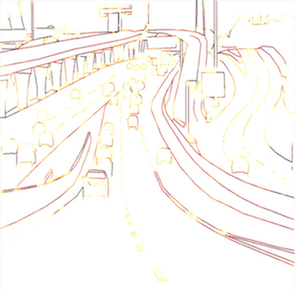}}\\
		&\fbox{\includegraphics[width=18mm,height = 12mm]{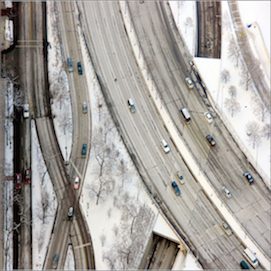}}&
		\fbox{\includegraphics[width=18mm,height = 12mm]{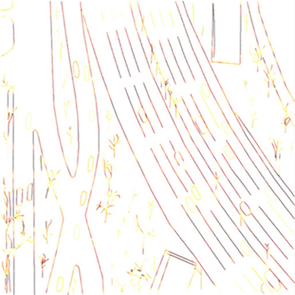}}&
		\fbox{\includegraphics[width=18mm,height = 12mm]{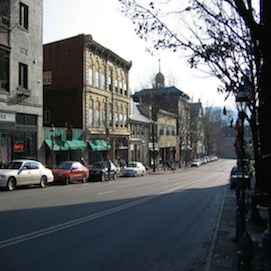}}&
		\fbox{\includegraphics[width=18mm,height = 12mm]{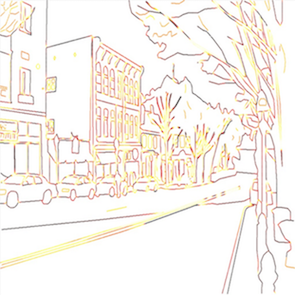}}&
		\fbox{\includegraphics[width=18mm,height = 12mm]{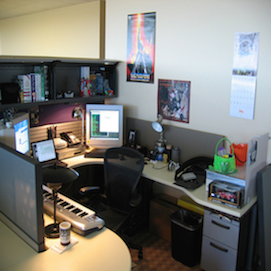}}&
		\fbox{\includegraphics[width=18mm,height = 12mm]{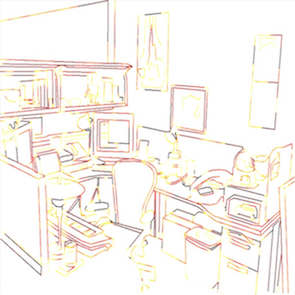}}\\\vspace{2mm}
		&\fbox{\includegraphics[width=18mm,height = 12mm]{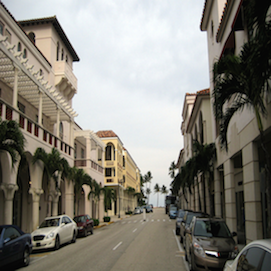}}&
		\fbox{\includegraphics[width=18mm,height = 12mm]{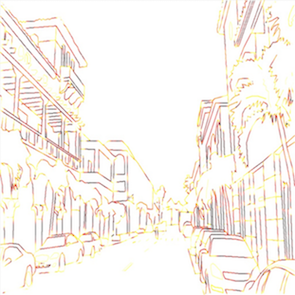}}&
		\fbox{\includegraphics[width=18mm,height = 12mm]{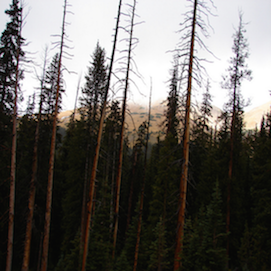}}&
		\fbox{\includegraphics[width=18mm,height = 12mm]{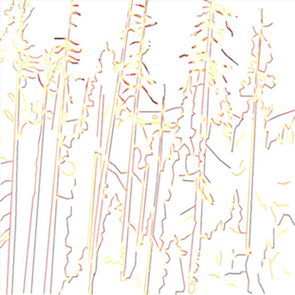}}&
		\fbox{\includegraphics[width=18mm,height = 12mm]{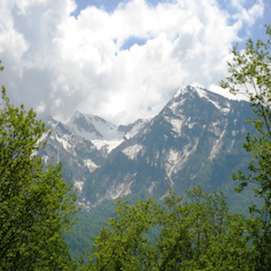}}&
		\fbox{\includegraphics[width=18mm,height = 12mm]{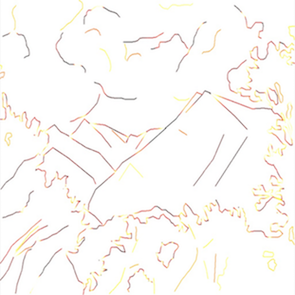}}\\
		
		& \multicolumn{2}{c}{\large Ribbon Symmetry} & \multicolumn{2}{c}{\large Separation} & \multicolumn{2}{c}{\large Taper Symmetry}\\

	\end{tabular}
	\caption{(Best viewed by zooming in on the PDF.) Examples of original photographs and the corresponding ribbon symmetry salience weighted, separation salience weighted and taper symmetry salience weighted scene contours, using a hot colormap to show increasing values. Whereas the Artist Scenes line drawings were produced by artists, the MIT67 and Places365  line drawings were machine-generated.}
	\label{fig:database}
\end{figure*}

\textbf{MIT67/Places365}
Given the limited number of scene categories in the Artist Scenes database, particularly for computer vision studies, we worked to extend our results to the two popular but much larger scene databases of photographs - MIT67 \cite{quattoni2009recognizing} (6700 images, 67 categories) and Places365 \cite{zhou2018places} (1.8 million images, 365 categories). Producing artist generated line drawings on databases of this size was not feasible, so instead we fine tuned the output of the Dollar edge detector \cite{DollarICCV13edges}, using the publicly available structured edge detection toolbox. From the edge map and its associated edge strength, we produced a binarized version, using per image adaptive thresholding. The binarized edge map was then processed to obtain contour fragments of width 1 pixel. Each contour fragment was then spatially smoothed by convolution of the coordinates of points along it, using a Gaussian of $\sigma = 1$, to mitigate discretization artifacts. The same parameters were used to produce all the MIT67 and Places365 line drawings. 
Figure \ref{fig:machine_generated} presents a comparison of a resultant machine-generated and an artist-generated line drawing for an office scene from the Artist Scenes database. Figure \ref{fig:database} shows several typical machine generated line drawings from the MIT67 and Places365 databases, but weighted by our ribbon symmetry salience measure.

\subsection{Computing Contour Salience}
\label{subsec:computing}
Computing contour salience for each line drawing required a number of steps. First, each connected region between scene contours was extracted. Second, we computed an AOF map for each of these connected components, as explained in Section \ref{sec:AOF}. For this we used a disk of radius 1 pixel, with 60 discrete sample points on it, to estimate the AOF integral. We used a threshold of  $\tau = 0.25$ on the AOF map, which corresponds to an object angle $\theta \approx 23$ degrees, to extract skeletal points. A typical example appears in Figure \ref{fig:overview} (top right). The resulting AOF skeleton was then partitioned into medial curves between branch points or between a branch point and an endpoint. We then computed a discrete version of each of the three salience measures in Section \ref{sec:salience}, within a interval $[\alpha,\beta]$ of length $2K+1$, centered at each medial axis point, with $K=5$ pixels.
Each scene contour point was then assigned the maximum of the two salience values at the closest points on the medial curves on either side of it, as illustrated in Figure \ref{fig:overview} (bottom middle and bottom right).

\subsection{Experiments on 50-50 Splits of Contour Scenes}

\begin{figure}[htpb]
	\centering
	\begin{tabular}{c@{\hskip15pt}c@{\hskip5pt}c}\vspace{1mm}
		\parbox[t]{2mm}{\rotatebox[origin=l]{90}{\Large\hspace{10mm}Ribbon}}&
		\fbox{\includegraphics[width=0.44\textwidth]{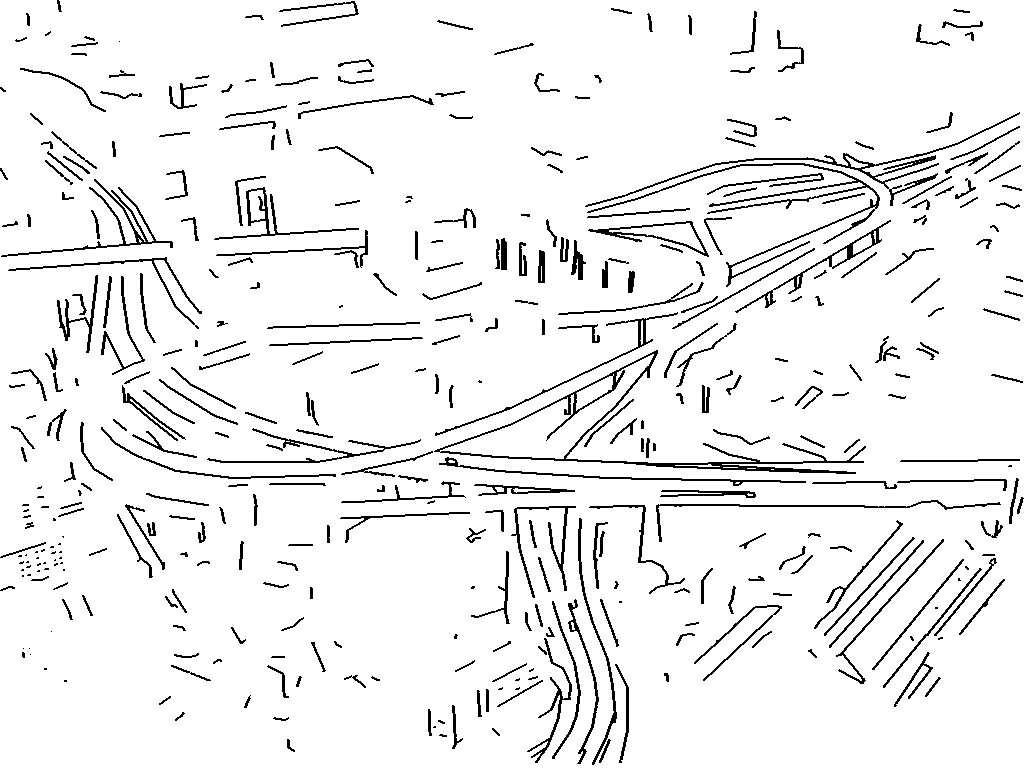}} &
		\fbox{\includegraphics[width=0.44\textwidth]{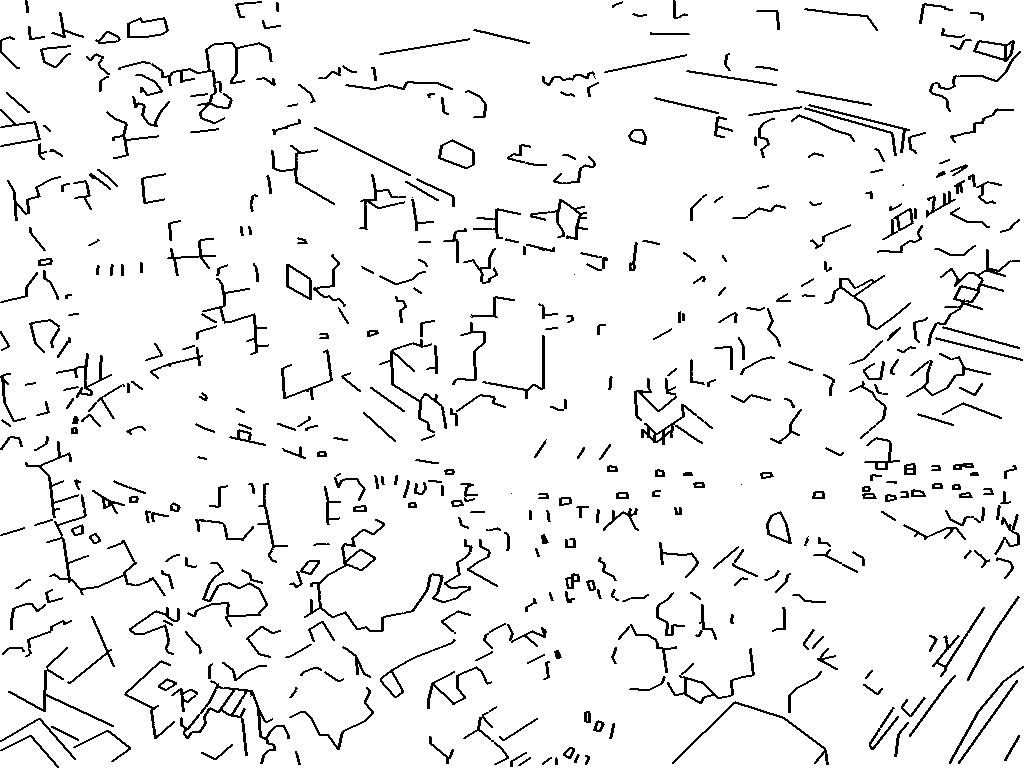}} \\\vspace{1mm}
		\parbox[t]{2mm}{\rotatebox[origin=l]{90}{\Large\hspace{10mm} Taper}}&
		\fbox{\includegraphics[width=0.44\textwidth]{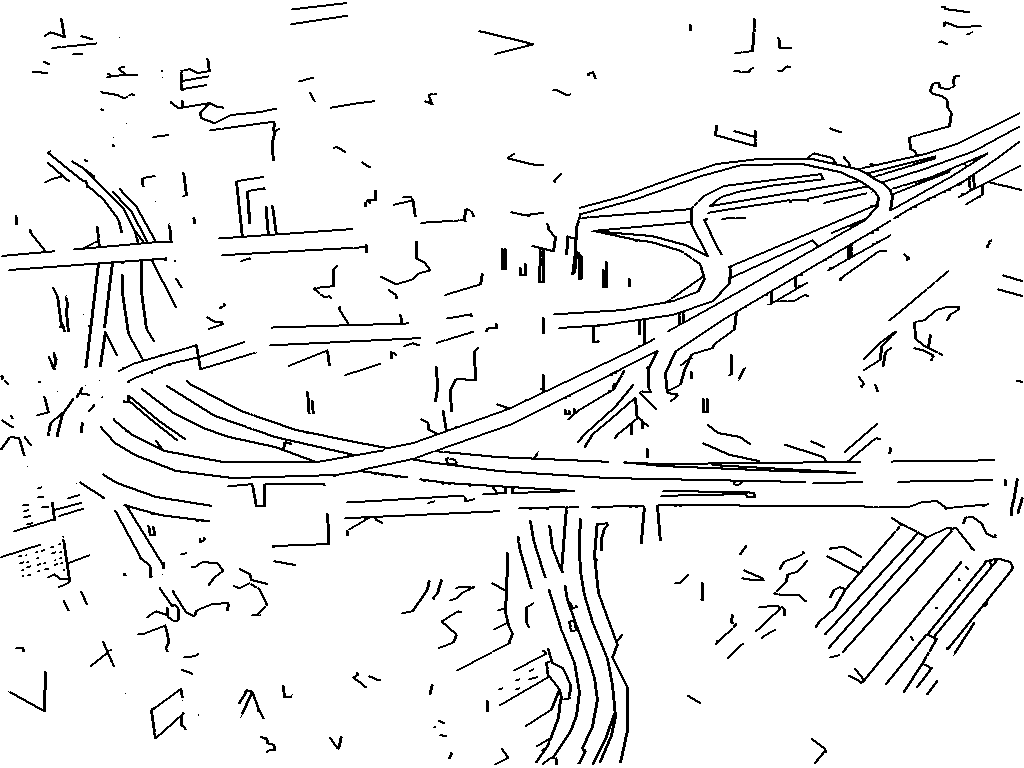}}& 
		\fbox{\includegraphics[width=0.44\textwidth]{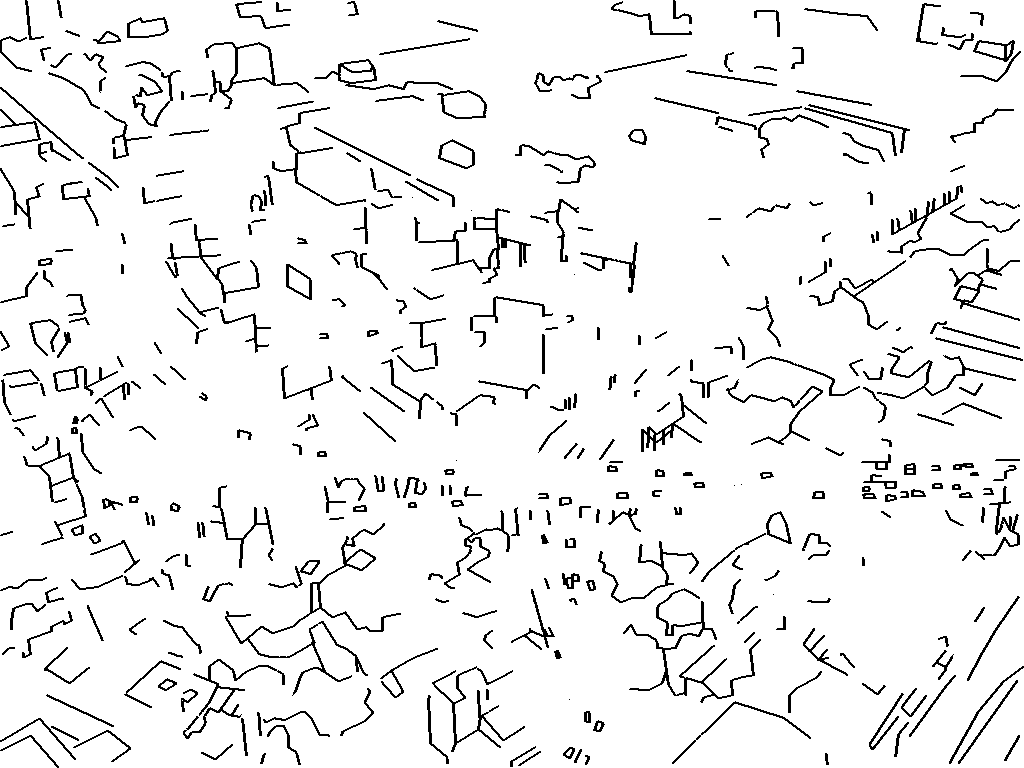}}\\
		\parbox[t]{2mm}{\rotatebox[origin=l]{90}{\Large\hspace{6mm} Separation}}&
		\fbox{\includegraphics[width=0.44\textwidth]{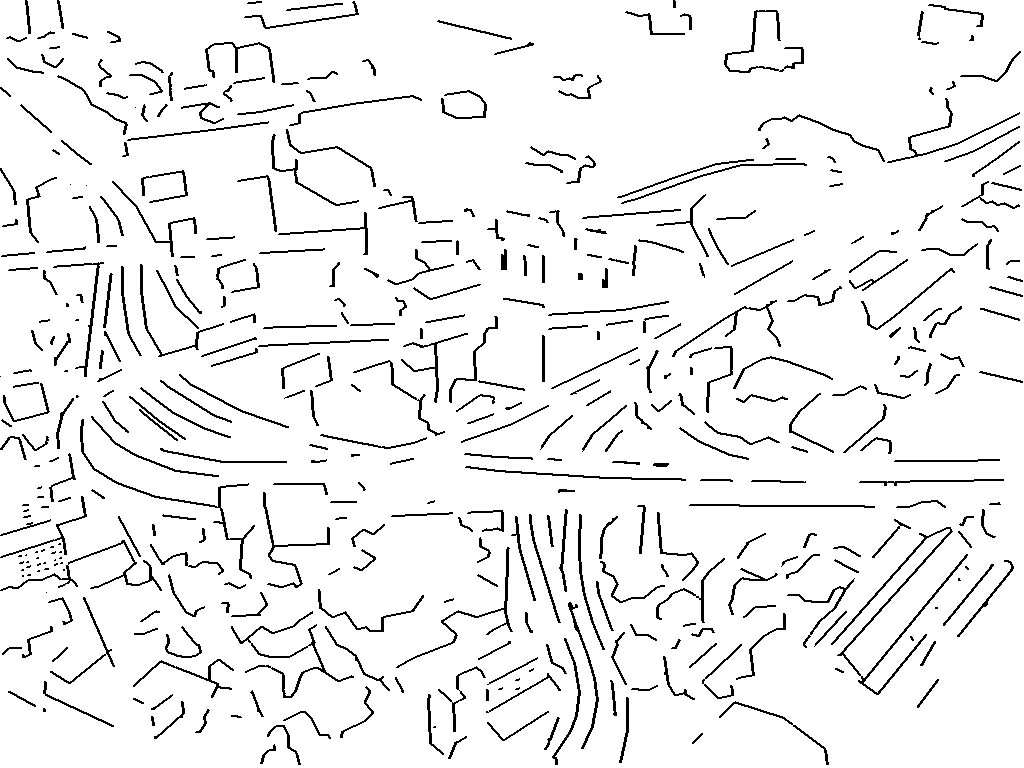}}& 
		\fbox{\includegraphics[width=0.44\textwidth]{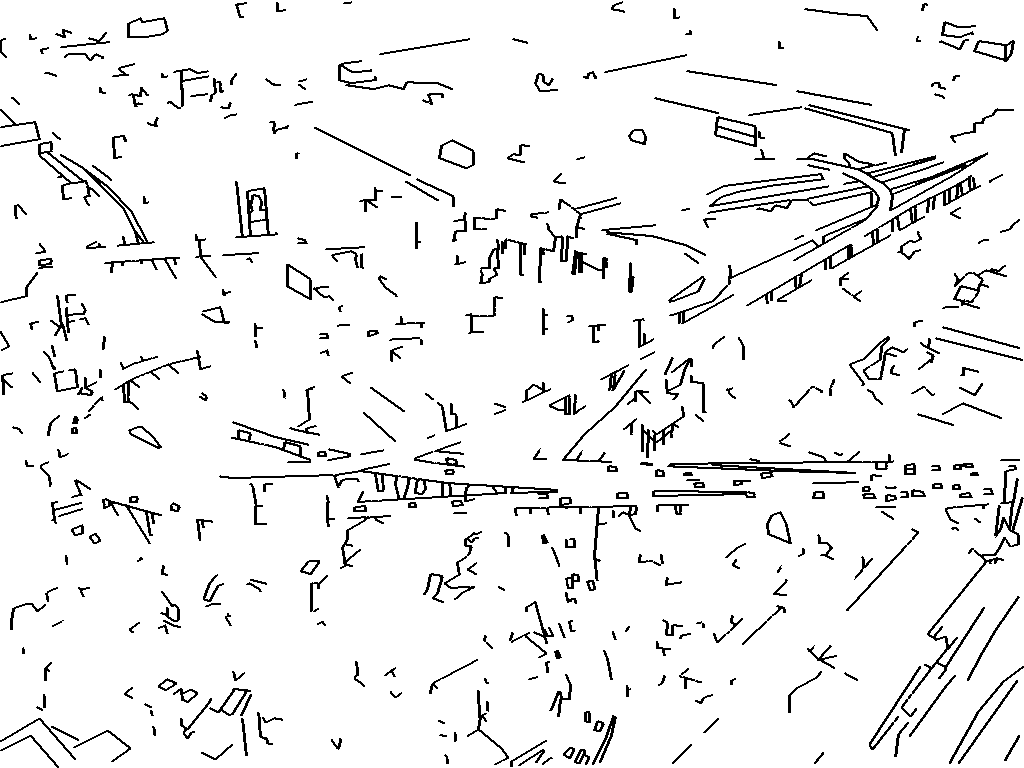}}\\
	\end{tabular}
	\caption{We consider the same highway scene as in Figure \ref{fig:overview} (top left) and create splits of the artist generated line drawings, each of which contains 50\% of the original pixels, based on ribbon symmetry (top row), taper symmetry (middle row) and local contour separation (bottom row) based salience measures. In each case the more salient half of the pixels is in the top row.}
	\label{fig:arc_length_scores}
\end{figure}

Our first set of experiments is motivated by recent work that shows that human observers benefit from cues such as contour symmetry in scene recognition from contours \cite{wilder2019local}. Our goal is to examine whether a CNN-based system also benefits from such perceptually motivated cues. Accordingly, we created splits of the top 50\% and the bottom 50\% of the contour pixels in each image of the Artist Scenes and MIT67 data sets, using the three salience measures, ribbon symmetry, taper symmetry and local contour separation. An example of the original intact line drawing and each of the three sets of splits is shown in Figure \ref{fig:arc_length_scores}, for the highway scene from the Artist Scenes dataset shown in Figure \ref{fig:overview}.

\begin{figure}[!h]
	\centering
	\includegraphics[width=0.6\textwidth]{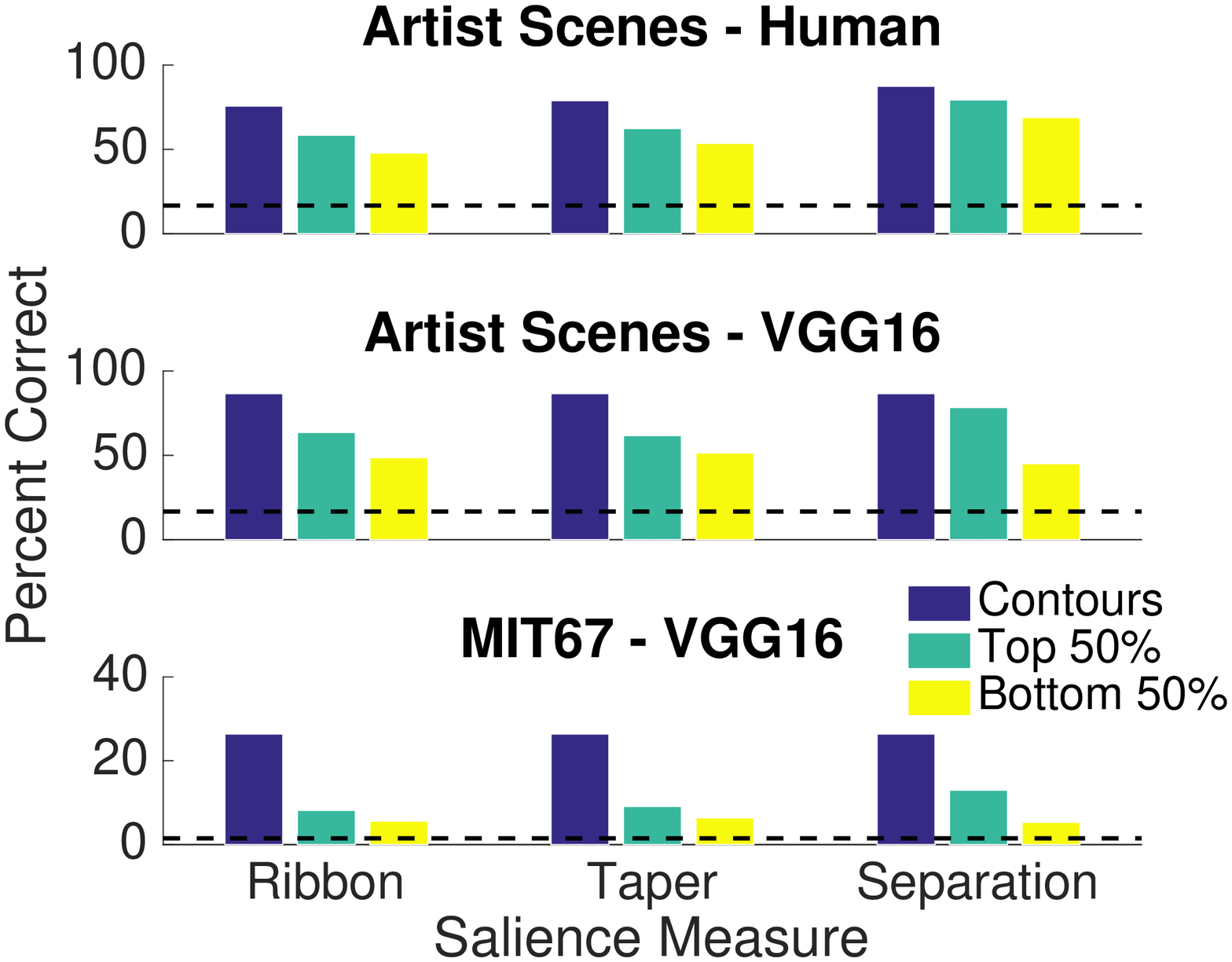}
	\caption{A comparison of human scene categorization performance (top row) with CNN performance (middle and bottom rows). As with the human observer data, CNNs perform better on the top 50\% half of each split according to each salience measure, that the bottom 50\% half. In each plot chance level performance (1/6 for Artist Scenes and 1/67 for MIT67) is shown with a dashed line.}
	\label{fig:table1}
\end{figure}

On the Artist Scenes dataset human observers were tasked with determining to which of six scene categories an exemplar belonged. The input was either the artist-generated line drawing or the top or the bottom half of a split by one of the salience measures. Images were presented for only 58 ms, and were followed by a perceptual mask, making the task difficult for observers, who would otherwise perform near 100\% correct. The results with these short image presentation durations, shown in Figure \ref{fig:table1} (top), demonstrate that human performance is consistently better with the top (more salient) half of each split than the bottom one, for each salience measure. The human performance is slightly boosted for all conditions in the separation splits, for which a different subject pool was used.

Carrying out CNN-based recognition on the Artist Scenes and MIT67 line drawing datasets presents the challenge that they are too small to train a large model, such as VGG-16, from scratch. To the best of our knowledge, no CNN-based scene categorization work has so far focused on line drawings of natural images. We therefore use CNNs that are pre-trained on RGB photographs for our experiments. For our experiments we use the VGG16 convolutional layer network architecture \cite{simonyan2014very} as well as the ResNet50 architecture \cite{he2016deep}. The network weights for VGG16 are those given in \cite{zhou2018places}, pre-trained on the ImageNet and Places365 datasets together (Hybrid1365-VGG), while for ResNet50 we used the original weights obtained by training on ImageNet. Following the methods in \cite{zhou2018places}, the images are processed by the Hybrid1365-VGG or ResNet50 network and the final fully connected layer is used as a feature vector input to an SVM classifier. 
The SVM classifier is trained on the feature maps of the line drawings. Classification is then performed on the held-out testing fold of the top 50\% and bottom 50\% split images for each salience measure. For all experiments on the Artist Scenes we use 5-fold cross validation. Top-1 classification accuracy is given, as a mean over the 5 folds, in Figure \ref{fig:table1} (middle). The CNN-based system mimics the trend we saw in human observers, namely that performance is consistently better for the top 50\% of each of the three splits. We interpret this as evidence that all three Gestalt motivated salience measures are beneficial for scene categorization in both computer and human vision.


For MIT67 we use the provided training/test splits and present the average results over 5 trials. The CNN-based categorization results are shown in Figure \ref{fig:table1} (bottom row). It is striking that even for this more challenging database, the CNN-based system still mimics the trend we saw in human observers, i.e., that performance is better on the top 50\% than on the bottom 50\% of each of the three splits and is well above chance. 


\subsection{Experiments With Salience Weighted Contours}
While we would expect that network performance would degrade when losing half the input pixels, the splits also reveal a significant bias in favor of our salience measures to support scene categorization. Can we exploit this bias to improve network performance when given the intact contours? To address this question, we carry out a second experiment where we explicitly encode salience measures for the CNN by feeding different features into the R-G-B color channels of the pre-trained network. We do this by using, in addition to the contour image channel, additional channels with same contours weighted by our proposed salience measures, each of which is in the interval $[0,1]$. All experiments again use a linear SVM classifier trained on the feature maps generated by the new feature-coded images.

\begin{figure}[!th]
	\begin{center}
		\includegraphics[width=0.5\textwidth]{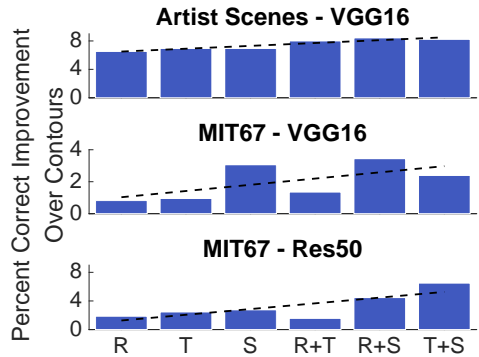}
		\caption{The increase in overall recognition accuracy by adding Ribbon (R), Taper (T) or Separation (S) salience weighted contour channels to the unweighted contour channel. All bars are above 0, showing that contour weighting by salience improves upon using just contours. The positive slope of the regression line shows that multiple weightings are generally better than a single weighting. }
		\label{fig:table2}
	\end{center}
\end{figure}

The results for the Artist Scenes dataset and for MIT67, are shown in Table \ref{tab:weighted}. It is apparent that with these salience weighted contour channels added, there is a consistent boost to the results obtained by using contours alone. We look at the performance gain in terms of the increase in percentage to top 1 recognition accuracy by adding one or more salience channels to the contour channel, in Figure \ref{fig:table2}.
In two out of the three cases the best performance boost comes from a combination of contours, ribbon symmetry salience and separation salience. We believe this is  because taper between local contours as a perceptual salience measure is very close in spirit to our ribbon salience measure. Local separation salience, on the other hand, provides a more distinct and complementary perceptual cue for grouping.

\begin{figure}[htpb]
	\begin{center}
		\includegraphics[width=0.5\textwidth]{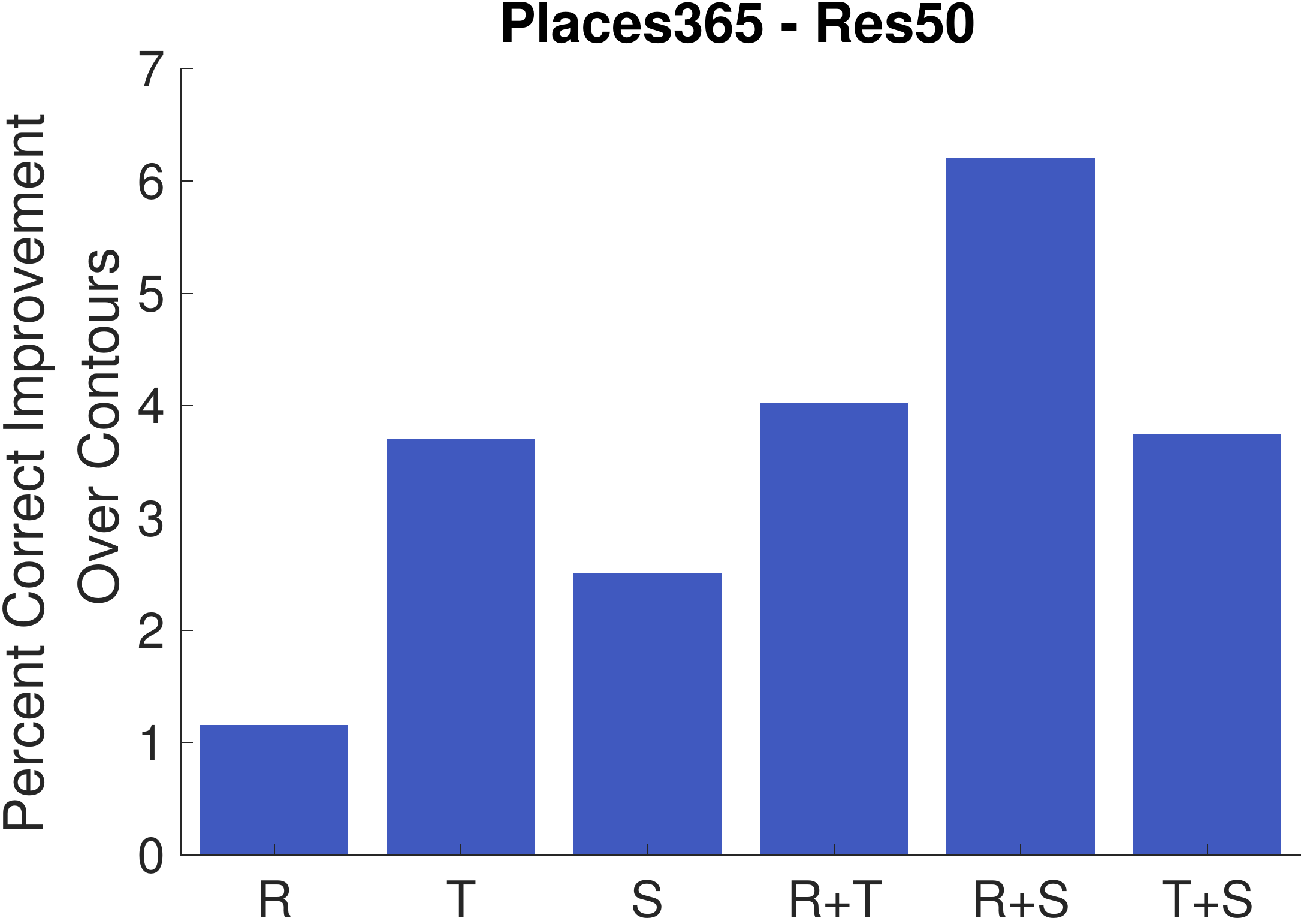}
		\caption{The increase in overall recognition accuracy by adding Ribbon (R), Taper (T) or Separation (S) salience weighted contour channels to the unweighted contour channel. R = Ribbon, T = Taper, S = Separation. All bars are above 0, showing that contour weighting by salience measures improves upon the performance of using just contours.}
		\label{fig:table3}
	\end{center}
\end{figure}

Encouraged by the above results, we repeated the same experiment for the much more challenging Places365 dataset, again using just a pre-trained network and a linear SVM. For this dataset chance recognition performance would be at 1/365 or 0.27\%. Our results are shown in Table \ref{tab:weightedplaces}, with the corresponding percentage increase in top 1 recognition accuracy by adding one or more salience measures to the contour channel, in Figure \ref{fig:table3}. Once again we see a clear and consistent trend of a benefit using salience weighted contours as additional feature channels to the contours themselves, with the best performance gain coming from the addition of ribbon symmetry salience and separation salience.

\begin{table}[!t]
	\begin{center}
		\begin{tabular}{|c|c|c|c|} 
			\hline
			\multirow{2}{*}{Channels} & Artist & \multicolumn{2}{|c|} {MIT67} \\\cline{2-4}
			&  VGG16 & VGG16 & Res50\\
			\hline
			Photos & 98.52& 72.52 & 66.19\\\hline
			Contours & 86.53 & 26.36 & 23.88\\\hline
			Contours, Ribbon & 93.05 & 27.18 & 25.74\\\hline
			Contours, Taper & 93.47 & 27.31 & 26.34\\\hline
			Contours, Separ. & 93.47 & 29.42 & 26.64\\\hline
			Contours, Ribbon, Taper & 94.53 & 27.71 & 25.44\\\hline
			Contours, Ribbon, Separ. & {\bf 94.96} & {\bf 29.80} & 28.35\\\hline
			Contours, Taper, Separ. & 94.75 & 28.75 & {\bf 30.37}\\\hline
			Ribbon, Taper, Separ. & 94.72 & 25.84 & 28.20\\\hline
		\end{tabular}
	\end{center}
	\caption{Top 1 performance in a 3-channel configuration, on Artist Scenes and MIT67, with an off-the-shelf pre-trained network and a linear SVM (see text). The top row shows the results of the traditional R,G,B input configuration, while the others show combinations of intact scene contours, and scene contours weighted by our salience measures.}
	\label{tab:weighted}
\end{table}

\begin{table}[t!]
	\begin{center}
		\begin{tabular}{|c|c|} 
			\hline
			Channels&  Places365 (Res50)\\
			\hline
			Photos & 33.04\\ \hline
			Contours & 8.02\\ \hline
			Contours, Ribbon & 9.18\\ \hline
			Contours, Taper & 11.73\\ \hline
			Contours, Separ & 10.53\\ \hline
			Contours, Ribbon, Taper & 12.05 \\ \hline
			Contours, Ribbon, Separ & {\bf 14.23} \\ \hline
			Contours, Taper, Separ & 11.77\\ \hline
			Ribbon, Taper, Separ & 12.64\\ \hline
		\end{tabular}
	\end{center}
	\caption{Top 1 performance in a 3-channel configuration on Places365, with an off-the-shelf pre-trained network and a linear SVM (see text). The top row shows the results of the traditional R,G,B input configuration, while the otheres show combinations of intact scene contours, and scene contours weighted by our salience measures.}
	\label{tab:weightedplaces}
\end{table}

\section{Conclusions}
We have reported the first study on CNN based recognition of complex natural scenes from line drawings derived from three databases of increasing complexity. To this end, we have demonstrated the clear benefit of using Gestalt motivated medial axis based salience measures, to weight scene contours according to their local ribbon and taper symmetry, and local contour separation. We hypothesize that making such contour salience weights explicit helps a deep network organize visual information to support categorization, in a manner which is not by default learned by these networks from the scene contour images alone. In our experiments we deliberately chose to use {\em off-the-shelf pretrained CNN models without any fine tuning}, to isolate the effect of these perceptually motivated scene contour grouping cues, and also the potential to perform scene categorization from contours alone, with color, shading and texture absent.
As such, our overall performance numbers are well below what they would have been had we done what is commonly done in benchmarking, namely, used data augmentation, hyper-parameter tuning and end-to-end fine tuning for each of our experiments. Those possibilities lie ahead, including an even more exciting one, which is the possibility to train a CNN model from scratch using our 1.8 million line drawings of Places365. The feasibility of fully trained networks on drawings has been demonstrated by work on free-hand sketches \cite{zou2018sketchyscene}, which, despite its superficial similarity with our work, follows a very different purpose. We plan on making our contour salience measure computation code, and our line drawing databases, publicly available.

\section*{Acknowledgments}
We are grateful to the Natural Sciences and Engineering Research Council of Canada (NSERC) for research funding.

\bibliographystyle{model1-num-names}
\bibliography{myrefs}

\end{document}